# Probing of Quantitative Values in Abstractive Summarization Models

Nathan M. White
James Cook University
`nathan.white1@jcu.edu.au`

## Abstract

Abstractive text summarization has recently become a popular approach, but data hallucination remains a serious problem, including with quantitative data. We propose a set of probing tests to evaluate the efficacy of abstract summarization models' modeling of quantitative values found in the input text. Our results show that in most cases, the encoders of recent SOTA-performing models struggle to provide embeddings that adequately represent quantitative values in the input compared to baselines, and in particular, they outperform random representations in some, but surprisingly not all, cases. Under our assumptions, this suggests that the encoder's performance contributes to the quantity hallucination problem. One model type in particular, DistilBART-CDM, was observed to underperform randomly initialized representations for several experiments, and performance versus BERT suggests that standard pretraining and finetuning approaches for the summarization task may play a role in underperformance for some encoders.

## 1 Introduction

Abstractive text summarization has become a popular summarization approach in recent years, especially with the advent of Transformer-based models (Vaswani et al., 2017) and the proliferation of large summarization datasets (Hermann et al., 2015; Narayan et al., 2018; cf. Zhao et al., 2020a). However, concomitant with the rise of abstractive summarization has been a phenomenon termed hallucination—i.e., the appearance of content in the generated summarization that is not in the original text and is typically erroneous (Nan et al., 2021). This represents a grave problem in the development of abstractive summarization; Maynez et al. (2020), for example, found that abstractive summarization systems produced summaries with inappropriate content more than 63% of the time.

One type of hallucination that plagues abstractive summarization in areas such as finance and economics is quantity hallucinations (Zhao et al., 2020a), where quantitative values appear in the summary output but not in the input.

Approaches have emerged to probing word embeddings to evaluate the effectiveness of quantitative representation (Naik et al., 2019; Wallace et al., 2019). A probing approach can be applied to the abstractive summarization context, given that for a model to summarize text, it must be able to recognize the quantitative values that typically need to be reproduced in the output.

We explore a number of novel and adapted quantitative probing tasks in the context of text summarization for the first time. We apply the probing tasks to encoder outputs to determine how adequately the quantities in the input are modeled.

We find that model architectures with recent state-of-the-art performance on the abstractive text summarization task struggled to adequately represent the quantitative values in their input as compared to baselines, suggesting that this is an important source of quantity hallucinations.

## 2 Rationale and Probing tasks

At a basic level, we assume that the problem of quantity hallucination when generating summaries, as in the examples in Table 1, derives from two possible contributors: 1) the encoder's representation of input, and 2) the decoder's generation of output. In turn, any contribution from the encoder is assumed to be due to the encoder's inadequate representation of the quantitative input. Under these assumptions, the encoder's contribution to the problem would be inversely proportional to the encoder output's performance on quantitative tasks.

We explore six numerical probing tasks to evaluate the encoder's output representations.

| Model | Hallucinated Numerical Values |
|---|---|
| Pegasus-XSum | Royal Bank of Scotland (RBS) has been fined **$5m (£3.2m)** by Hong Kong regulators for failing to properly monitor its traders. |
| T5-CDM | Sandra Bland, 28, was arrested during a traffic stop on Texas Highway **88**. |

Table 1. Examples of hallucinated numerical values from abstractive summaries generated by selected models under consideration, based on the XSum dataset (Narayan et al., 2018). Hallucinated values are in bold.

Each task has been selected due to its real-world significance to financial news and the resulting potential for involvement in text summarization. The focus of each task is to discern whether the signal of the quantity or unit is recoverable from the summarization encoder's output.

**Percent Decoding** This task adapts the Decoding task of Wallace et al. (2019) to single-decimal float values indicating percents by casting input numbers as strings encoding percents in the format "10.3%", where the output should regress to the real-world float representation of the value, e.g., 0.103. The probing network receives the concatenated output of the model encoder and consists of three fully connected layers with ReLU activations for the first two layers. Mean squared error serves as the training loss.

**Basis Point Decoding** This task is similar to percent decoding, though with inputs representing basis points, a ubiquitous quantitative form in financial media representing hundredths of percents, such as "15 basis points" for 0.0015. The same network is used as in percent decoding.

**Order Decoding** This task considers float values with text representing multiples of 10, e.g., "15.3 billion". This task seeks to regress to the natural log value of the input value, e.g., "15.3 billion" as 10.1847. The probing network used is the same as with percent decoding.

**Ranges** This task considers the endpoints of ranges, such as "27.3-65.1". The task here is to regress to each of the two values as a joint task. For this, we consider a network with two outputs. The probing model is trained using a combined mean squared error loss calculated from each of the two outputs and added.

**Addition** This task adapts the Addition task of Wallace et al. (2019) to float values, and uses the same architecture as the Percent Decoding task. Training uses mean squared error loss, with a single float value representing the sum as target.

**Unit Identification** This task is cast as a multiclass classification over a set of 173 unit types associated with numerical values found in the financial news datasets of Malo et al. (2014) and Turenne et al. (2021). This task probes the semantic representations of units associated with numerals. Randomly sampled synthetic data points containing sequences of a single-decimal float value and a unit, e.g., "14.3 hours", serve as input. The embeddings are fed into a BiLSTM layer, with the hidden layer result concatenated and fed into a fully connected layer with a softmax activation. Categorical cross-entropy is the loss.

## 3 Text summarization models

Eight Transformer-based models are considered for the probing tasks. Several have achieved state-of-the-art (SOTA) results for the abstractive text summarization task. The models are:

- **Pegasus-XSum** (Zhang et al., 2020), using Pegasus$_{BASE}$, as fine-tuned on the XSum dataset (Narayan et al., 2018);
- **Pegasus-CDM**, the equivalent fine-tuned on the CNN/Daily Mail dataset (Nallapati et al., 2016);
- **T5-CDM**, using T5$_{BASE}$ (Raffel et al., 2020), with fine-tuning on CNN/Daily Mail;
- **BART-XSum**, i.e., BART$_{Large}$ (Lewis et al., 2020) with fine-tuning on XSum;
- **BART-CDM**, fine-tuned on the CNN/Daily Mail dataset;
- **DistilBART-XSum** (Shleifer and Rush, 2020), using DistilBART-12-6, fine-tuned on XSum;
- **DistilBART-CDM**, fine-tuned on CNN/Daily Mail; and
- **ProphetNet-CDM** (Qi et al., 2020), fine-tuned on CNN/Daily Mail.

Three baselines are also considered:

- **Random Vectors** (following Wallace et al., 2019), a random representation which allows comparison of Transformer-based models' representations with chance;
- **BERT** (Devlin et al., 2019) Untrained, using BERT$_{BASE}$, a Transformer-based

|  | Percents (RMSE) | | Basis Points (RMSE) | | Orders (RMSE Log scale) | | Ranges (RMSE) | |
|---|---|---|---|---|---|---|---|---|
|  | [0.0, 99.9] | [0.0, 999.9] | [0.0, 99.9] | [0.0, 999.9] | [0.0, 99.9] | [0.0, 999.9] | [0.0, 99.9] | [0.0, 999.9] |
| Baselines: | | | | | | | | |
| Random Vectors | 9.365 | 146.445 | 1.629 | 2.114 | 607.507 | 191.655 | 2652.395 | 23607.775 |
| BERT$_{Untrained}$ | 9.464 | 85.727 | 0.090 | 1.075 | 141.375 | 148.243 | 1130.263 | 12690.523 |
| BERT$_{Trained}$ | 5.617 | **60.521** | 0.309 | 0.786 | 233.457 | **46.643** | **982.246** | 17108.312 |
| Pegasus-XSum | *9.756* | 94.499 | 0.111 | 0.105 | **128.907** | 95.481 | 1465.958 | 12611.591 |
| Pegasus-CDM | *11.349* | 121.388 | 0.170 | **0.078** | 241.866 | 141.945 | 2113.631 | 12010.757 |
| T5-CDM | *11.711* | *147.866* | 0.134 | 0.556 | 141.658 | 117.020 | 1419.279 | 18874.290 |
| BART-XSum | 5.332 | 78.992 | 0.196 | 1.117 | 313.358 | *268.685* | 1477.931 | 14732.467 |
| BART-CDM | 9.419 | 113.235 | 0.295 | 0.731 | 236.019 | *226.570* | 1488.847 | 13105.150 |
| DistilBART-XSum | 6.234 | 67.412 | 0.303 | 0.864 | 308.755 | *237.245* | 1440.513 | 14864.469 |
| DistilBART-CDM | 6.209 | 111.583 | *0.579* | 0.715 | 306.743 | *231.654* | 1483.148 | 15613.766 |
| ProphetNet-CDM | **4.601** | 70.252 | **0.083** | 1.047 | 144.047 | 179.836 | 1247.854 | **9561.739** |

Table 2: Mean results for randomized experiments for the three Decoding tasks and the Ranges task with floats in ranges [0.0, 99.9] and [0.0, 999.9], with mean results worse than all baselines italicized.

architecture with randomly initiated weights, which represents the Transformer architecture's contribution without dataset-specific training and so allows for measuring the contribution of the training approach; and

- **BERT** Trained, using BERT$_{BASE}$, which lacks training for a summarization objective and so allows the measurement of the objective's contribution to performance.

## 4 Experiments and results

We consider each task presented in Section 2 above. Aspects of the training and evaluation approach are found in Section 4.1 below, with additional details provided in Appendix A; results appear in Section 4.2.

|  | Addition (RMSE) | | Unit ID (Accuracy) | |
|---|---|---|---|---|
|  | [0.0, 99.9] | [0.0, 999.9] | [0.0, 99.9] | [0.0, 999.9] |
| Baselines: | | | | |
| Random Vectors | 3309.333 | 34484.099 | 0.995 | **0.995** |
| BERT$_{Untrained}$ | 1665.166 | 15140.284 | 0.989 | 0.994 |
| BERT$_{Trained}$ | 1904.377 | 17845.853 | 0.995 | 0.994 |
| Pegasus-XSum | 2416.222 | 13410.940 | 0.994 | 0.994 |
| Pegasus-CDM | 3038.839 | 20372.009 | 0.995 | *0.993* |
| T5-CDM | 2015.641 | 24095.699 | 0.992 | **0.995** |
| BART-XSum | 2147.291 | 15790.598 | 0.995 | *0.725* |
| BART-CDM | 1745.232 | 13550.998 | 0.991 | *0.443* |
| DistilBART-XSum | 1537.111 | **11799.463** | **0.996** | 0.994 |
| DistilBART-CDM | 1574.491 | 13100.640 | **0.996** | *0.433* |
| ProphetNet-CDM | **1524.449** | 15897.303 | *0.985* | *0.989* |

Table 3: Mean results for Addition and Unit Identification tasks.

### 4.1 Training and evaluation

Each of the probing tasks are applied using an interpolation setting (cf. Wallace et al., 2019), selecting ranges of float values of [0.0, 99.9] and [0.0, 999.9]. Five runs were conducted for each experiment, which involved randomly generated datasets of 10,000 training data points and 1000 test data points.

The experiment models were composed of two parts: the fine-tuned Transformer-based model with its parameters frozen and a probing model trained in each experiment. Grid searches determined the best learning rates for each task were between 0.3 and 1e-06, with most values at or above 3e-04, and each task was run for 1000 epochs with early stopping after 20 epochs.

For results, accuracy is provided for Unit Identification, and Log-scaled Root Mean Squared Error (RMSE) is used with Order Decoding. For the other tasks, simple RMSE is provided. The mean for each task is provided for five runs with newly sampled data for each run, with standard deviations provided in Appendix B.

### 4.2 Results

Our results (Tables 2 and 3) show that the encoders of all of the abstractive summarization models considered provide embeddings that model the numerical values to some degree as compared to the Random Vectors baseline in most tasks. Moreover, when considering the mean results, the following observations can be made.

**Random Vectors sometimes outperform** The encoder outputs from several summarization models provide representations for quantities that are, surprisingly, worse than chance, as represented by the Random Vectors. This is especially true of DistilBART-CDM.

**BERT often outperforms** $BERT_{Trained}$ performs the best on three experiments, and outperforms T5-CDM on eight out of 12 tasks. $BERT_{Untrained}$ outperforms all trained models besides $BERT_{Trained}$ on one task, and BART-XSum in particular on nine tasks. Given the relatively poor performance of fine-tuned models against the baselines, this suggests that standard pretraining and/or fine-tuning methods may play a role in inferior modeling of quantitative values, especially where they underperform against even a randomly initialized model such as $BERT_{Untrained}$.

**No model provides consistent superior performance** This suggests that existing methods for text summarization do not provide a decisive advantage to adequately represent quantitative values, though in most cases, Transformer-based representations provide an advantage over Random Vectors.

Altogether, the summarization model encoders produce representations of the input that struggle to consistently outperform baselines. Since these representations serve as intermediate inputs into text summarization decoders, this would present a problem for any summarization task. If the encoder's output does not adequately represent numerical values, then the decoder would struggle to reconstruct these in training. This would be expected to lead the decoder to generate numerical values not necessarily represented in the encoder's output to compensate to match the target. While the confirmation of this is left for further research, the lack of superior performance in text summarization encoders' modeling of quantitative values is clear.

## 5 Related work

### 5.1 Abstractive summarization and hallucination

Abstractive summarization has exhibited a degree of hallucinated facts in its generated summaries (Cao et al., 2018; Falke et al., 2019; Maynez et al., 2020; Zhao et al., 2020a). Given the limitations of ROUGE or BLEU when considering these factual inaccuracies (Kryściński et al., 2019), a number of metrics have been devised to attempt to capture factual inaccuracies or other inconsistencies, including with model-based methods (Goodrich et al., 2019; Kryściński et al., 2020; Vasilyev et al., 2020; Mishra et al., 2021; Zhou et al., 2021a; Laban et al., 2022), including named entity recognition-based (Nan et al., 2021) and question-answering-based approaches (Durmus et al., 2020; Wang et al., 2020). Pagnoni et al. (2021) propose a typology of factual errors in summarization models and evaluates a range of factuality metrics, while Gabriel et al. (2021) consider a meta-evaluation framework for metrics.

### 5.2 Probing models

Probing of language models has focused on semantics (Yaghoobzadeh et al., 2019; Zhao et al., 2020b), morphosyntactic structure (Linzen et al., 2016; Bacon & Regier, 2018; Jawahar et al., 2019; McCoy et al., 2019; Lepori & McCoy, 2020), linguistic capability (Liu et al., 2019), and knowledge of context and surface properties (Adi et al., 2017; Khandelwal et al., 2018). Several studies have explored a combination of these areas (Conneau et al., 2018; Lin et al., 2019; Tenney et al., 2019; Wieting & Kiela, 2019; Mosbach et al., 2020; Puccetti et al., 2021). Others have considered sentiment analysis (Perone et al., 2018) or named entity recognition (Jin et al., 2019). A few have considered numeracy probing (Naik et al., 2019; Wallace et al., 2019).

## 6 Conclusion

We consider a set of probing tests to evaluate the efficacy of abstractive summarization models' modeling of quantitative values found in the input text. Our results show that in most cases, the encoders of recent SOTA-performing models struggle to provide embeddings that adequately represent quantitative values in the input compared to baselines, and in particular, they outperform random representations in some, but surprisingly not all, cases. Under our assumptions, this suggests that the encoder's performance contributes to the quantity hallucination problem. One model type in particular, DistilBART-CDM, was observed to underperform randomly initialized representations for several experiments, and performance versus BERT suggests that standard pretraining and fine-tuning approaches for the abstractive summarization task may play a role in underperformance for some encoders.

## Appendix A. Supplementary Details for Training and Evaluation

Training and evaluation utilized the PyTorch (Paszke et al., 2019) and Transformers (Wolf et al., 2020) libraries, versions 1.10.2 and 4.17.0, respectively. PyTorch is available under a BSD license and Transformers under an Apache 2.0 license. Pretrained models were accessed via Transformers. Training and evaluation were conducted via an infrastructure using one Tesla V100 GPU in each experiment; compute time for each run ranged from about 3 minutes to 2 hours, 45 minutes, depending on early stopping and the task.

Numpy (Harris et al., 2020) and Scikit-Learn (Pedregosa et al., 2011) were used as part of dataset generation; Numpy is available under a liberal BSD license and Scikit-Learn under a permissive simplified BSD license.

In addition to the basic hyperparameter details as considered in section 4.1 above, exploratory testing determined hyperparameters for batch size at 128, a maximum gradient norm of 5, and best momentum rates typically were either 0.5 or 0.7. In addition, extensive hyperparameter grid searches showed that each combination of task and model architecture type required its own settings for learning rate and momentum; momentum was set to 0.5 or 0.7 depending on the experiment based on the results of the grid searches.

The number of parameters used for each Transformer-based model type is as specified in their original papers and as determined by the model configurations in Transformers. The Random Vectors embedding was 30,522 x 768, representing the vocabulary size for BERT and the size of the embedding. For the probing task model components, the three Decoding tasks used a hidden dimension of 100 as did Addition, while the Unit Identification task used a BiLSTM layer with a hidden dimension of 5, and the Ranges task used a hidden dimension of 50.

The weights of the text summarization or baseline model under consideration were frozen for both training and evaluation, while the probing component was trained. The frozen, pretrained models used were the conditional generation versions of the models as published in Transformers (Wolf et al., 2020), as the appropriate architectures for abstractive text summarization.

Torchmetrics (Detlefsen et al., 2022) was used to calculate accuracy for the Unit Identification task, while PyTorch was used for the other metrics.

## Appendix B. Mean with Standard Deviations of Experiment Results

The mean values of experiment results with their corresponding standard deviations from the Percent and Basis Point Decoding tasks appear in Table 4, while those from the Order Decoding and Ranges tasks appear in Table 5, and those from the Addition and Unit Identification tasks appear in Table 6.

|  | Percents (RMSE) | | Basis Points (RMSE) | |
| --- | --- | --- | --- | --- |
|  | [0.0, 99.9] | [0.0, 999.9] | [0.0, 99.9] | [0.0, 999.9] |
| Baselines: | | | | |
| Random Vectors | 9.365±0.778 | 146.445±5.379 | 1.629±2.629 | 2.114±1.362 |
| BERT$_{Untrained}$ | 9.464±0.544 | 85.727±14.328 | 0.090±0.012 | 1.075±0.108 |
| BERT$_{Trained}$ | 5.617±3.030 | **60.521**±29.989 | 0.309±0.277 | 0.786±0.392 |
| Pegasus-XSum | *9.756*±1.017 | 94.499±48.830 | 0.111±0.078 | 0.105±0.052 |
| Pegasus-CDM | *11.349*±0.898 | 121.388±15.151 | 0.170±0.092 | **0.078**±0.027 |
| T5-CDM | *11.711*±1.559 | *147.866*±8.899 | 0.134±0.057 | 0.556±0.486 |
| BART-XSum | 5.332±5.336 | 78.992±41.528 | 0.196±0.270 | 1.117±0.169 |
| BART-CDM | 9.419±1.640 | 113.235±0.882 | 0.295±0.388 | 0.731±0.593 |
| DistilBART-XSum | 6.234±5.153 | 67.412±58.821 | 0.303±0.394 | 0.864±0.467 |
| DistilBART-CDM | 6.209±4.961 | 111.583±6.914 | *0.579*±0.438 | 0.715±0.579 |
| ProphetNet-CDM | **4.601**±2.258 | 70.252±30.246 | **0.083**±0.015 | 1.047±0.108 |

Table 4: Mean and standard deviation of results from Percent and Basis Point Decoding tasks.

|  | Orders (RMSE Log scale) | | Ranges (RMSE) | |
| --- | --- | --- | --- | --- |
|  | [0.0, 99.9] | [0.0, 999.9] | [0.0, 99.9] | [0.0, 999.9] |
| Baselines: | | | | |
| Random Vectors | 607.507±13.347 | 191.655±7.714 | 2652.395±228.999 | 23607.775±1207.387 |
| BERT$_{Untrained}$ | 141.375±14.092 | 148.243±19.873 | 1130.263±67.715 | 12690.523±828.718 |
| BERT$_{Trained}$ | 233.457±15.627 | **46.643**±30.940 | **982.246**±62.832 | 17108.312±1564.177 |
| Pegasus-XSum | **128.907**±82.565 | 95.481±81.994 | 1465.958±216.848 | 12611.591±733.364 |
| Pegasus-CDM | 241.866±12.561 | 141.945±123.175 | 2113.631±157.748 | 12010.757±6489.031 |
| T5-CDM | 141.658±33.118 | 117.020±86.841 | 1419.279±176.572 | 18874.290±2091.475 |
| BART-XSum | 313.358±22.206 | 268.685±28.576 | 1477.931±595.316 | 14732.467±1615.047 |
| BART-CDM | 236.019±57.062 | 226.570±23.537 | 1488.847±442.763 | 13105.150±6998.130 |
| DistilBART-XSum | 308.755±15.428 | 237.245±10.480 | 1440.513±318.256 | 14864.469±2257.984 |
| DistilBART-CDM | 306.743±7.073 | 231.654±4.191 | 1483.148±390.091 | 15613.766±736.186 |
| ProphetNet-CDM | 144.047±10.252 | 179.836±39.887 | 1247.854±390.352 | **9561.739**±4480.605 |

Table 5: Mean and standard deviation of results from Order Decoding and Ranges tasks.

|  | Addition (RMSE) | | Unit ID (Accuracy) | |
|---|---|---|---|---|
|  | [0.0, 99.9] | [0.0, 999.9] | [0.0, 99.9] | [0.0, 999.9] |
| Baselines: | | | | |
| Random Vectors | 3309.333±563.693 | 34484.099±1110.961 | 0.995±0.002 | **0.995**±0.002 |
| BERT$_{Untrained}$ | 1665.166±95.971 | 15140.284±2492.251 | 0.989±0.016 | 0.994±0.003 |
| BERT$_{Trained}$ | 1904.377±289.697 | 17845.853±2576.767 | 0.995±0.002 | 0.994±0.008 |
| Pegasus-XSum | 2416.222±244.690 | 13410.940±5313.597 | 0.994±0.002 | 0.994±0.003 |
| Pegasus-CDM | 3038.839±330.205 | 20372.009±1952.716 | 0.995±0.003 | *0.993*±0.001 |
| T5-CDM | 2015.641±170.722 | 24095.699±3530.997 | 0.992±0.006 | **0.995**±0.001 |
| BART-XSum | 2147.291±448.180 | 15790.598±1954.166 | 0.995±0.003 | *0.725*±0.371 |
| BART-CDM | 1745.232±106.215 | 13550.998±442.790 | 0.991±0.006 | *0.443*±0.504 |
| DistilBART-XSum | 1537.111±237.554 | **11799.463**±6016.980 | **0.996**±0.001 | 0.994±0.004 |
| DistilBART-CDM | 1574.491±285.660 | 13100.640±6207.451 | **0.996**±0.003 | *0.433*±0.510 |
| ProphetNet-CDM | **1524.449**±409.306 | 15897.303±6698.941 | *0.985*±0.005 | *0.989*±0.002 |

Table 6: Mean and standard deviation of results from Addition and Unit Identification tasks.